\pdfoutput=1
\documentclass[11pt]{article}

\usepackage{ACL2023}
\usepackage{times}
\usepackage{latexsym}

\usepackage[T1]{fontenc}
\usepackage[utf8]{inputenc}

\usepackage{microtype}

\usepackage{inconsolata}

\usepackage{booktabs}
\usepackage{graphicx}
\usepackage{multirow}
\usepackage{floatrow}
\usepackage{amssymb}

\definecolor{purple}{HTML}{BE2BBB}
\definecolor{gray}{HTML}{595454}
\newcommand{\eg}{\emph{e.g.}, }

\newcommand{\comment}[1]{}

\title{Language Models Are Few-shot Learners for Prognostic Prediction}

 \author{Zekai Chen \and Mariann Micsinai Balan \and Kevin Brown \\
   Bristol-Myers Squibb, NJ, USA \\
   \texttt{\{zekai.chen\}@bms.com} \\
   }
\begin{document}
\maketitle
 \begin{abstract}
Clinical prediction is an essential task in the healthcare industry. However, the recent success of transformers, on which large language models are built, has not been extended to this domain. In this research, we explore the use of transformers and language models in prognostic prediction for immunotherapy using real-world patients' clinical data and molecular profiles. This paper investigates the potential of transformers to improve clinical prediction compared to conventional machine learning approaches and addresses the challenge of few-shot learning in predicting rare disease areas. The study benchmarks the efficacy of baselines and language models on prognostic prediction across multiple cancer types and investigates the impact of different pretrained language models under few-shot regimes. The results demonstrate significant improvements in accuracy and highlight the potential of NLP in clinical research to improve early detection and intervention for different diseases.
\end{abstract}

\section{Introduction}
\label{sec:intro}

Predicting and measuring treatment response is among the most fundamental tasks in clinical medicine. Particularly, in cancer immunotherapy~\citep{Pardoll2012TheBO}, antibodies against programmed death-1/programmed death ligand 1 (PD-1/PD-L1) have led to US FDA approval of several PD-1/PD-L1 treatment strategies for patients with metastatic cancer. However, not all patients derive clinical benefits ~\citep{Topalian2016MechanismdrivenBT}, emphasizing the need to identify who will respond to immunotherapy~\citep{Chowell2021ImprovedPO}. Thus, accurate treatment response and disease progress forecast based on the patient's clinical features and molecular profile will effectively improve the treatment efficiency and spur the development of precise medication. In order to facilitate medical decision-making and health outcomes, clinical prediction models~\citep{Steyerberg2008ClinicalPM,vanSmeden2021ClinicalPM} play an increasingly crucial role in contemporary clinical care by informing professionals, patients, and their relatives about outcome risks. 

\begin{figure}[!t]	
\centering
\includegraphics[width=0.9\linewidth]{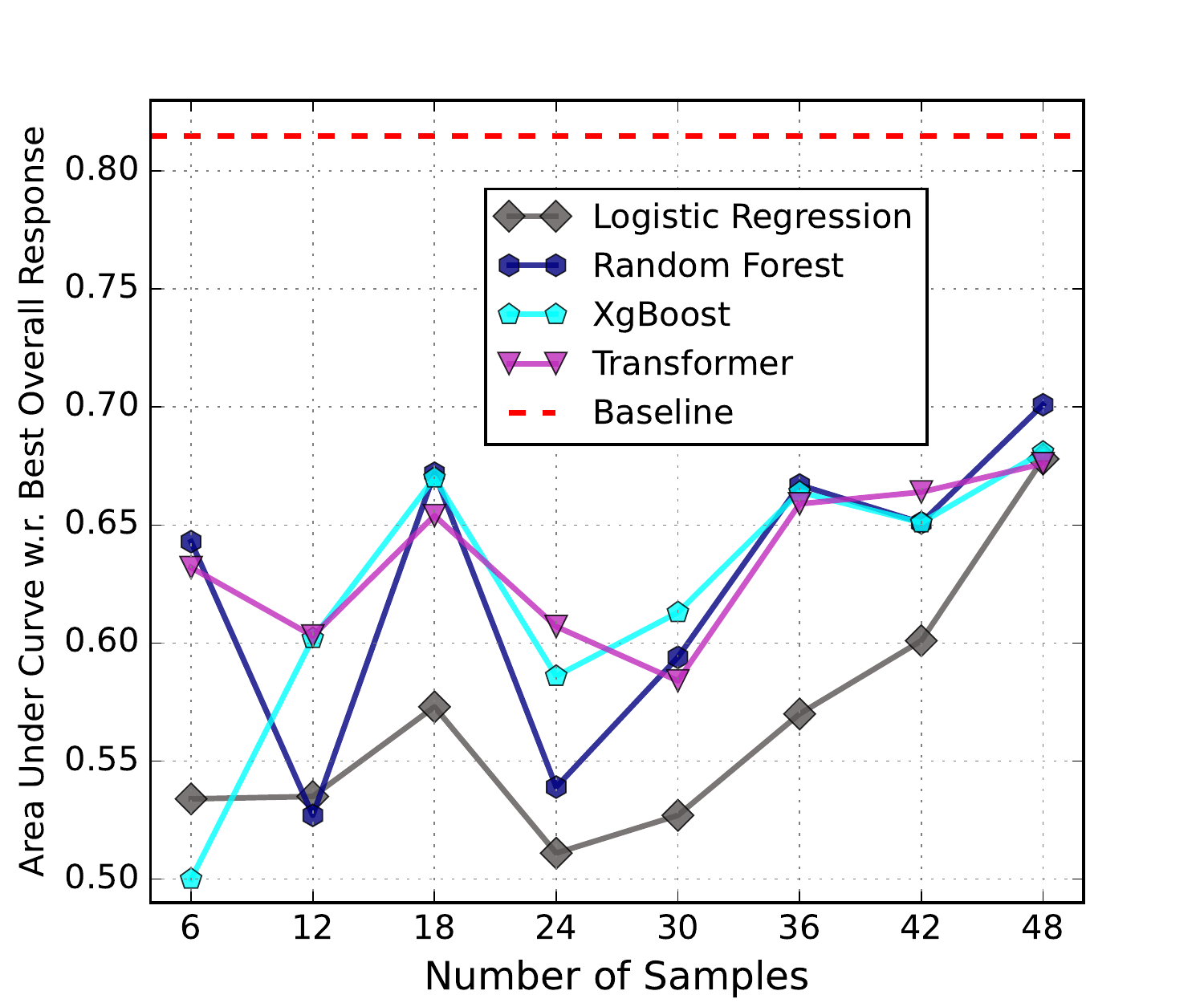}
\caption{\textbf{Pilot study}. We evaluate the prediction performance (\textbf{AUC}) of a patient's probability of immunotherapy response  across multiple cancer types under settings with a small number of training samples on a public clinical dataset from~\citet{Chowell2021ImprovedPO}.}
%\vspace{-10pt}
\label{fig:intro}
\end{figure}

\begin{figure*}
\centering
\includegraphics[width=\linewidth]{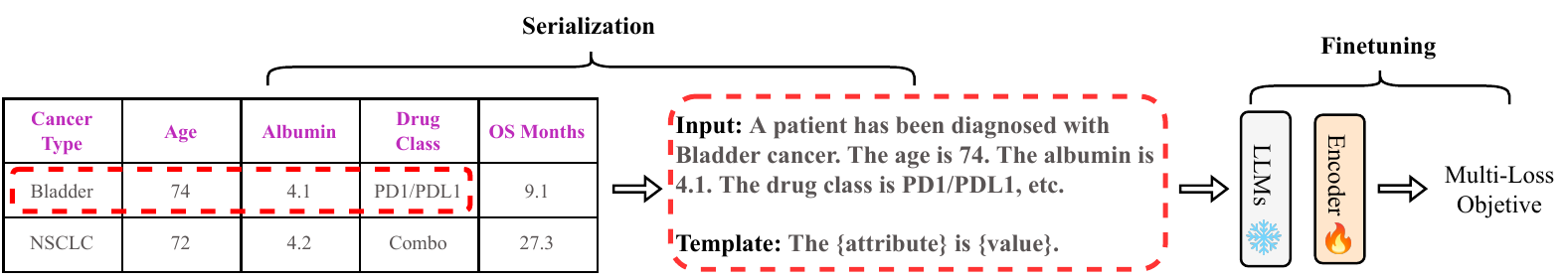}
\caption{\textbf{An illustration of adapting LLMs for clinical prediction.} The clinical data entry is first serialized into sequences of natural language tokens and then fed into the \emph{frozen} LLMs, followed by a randomly initialized encoder (transformers or MLPs or identical blocks) to finetune with the multi-loss objective same as Eq.~\ref{eq:loss}.}
%\vspace{-10pt}
\label{fig:llms}
\end{figure*}

Given the fact that most clinical data is stored in tabular form, current mainstream machine learning approaches~\citep{Topol2019HighperformanceMT,Rajkomar2019MachineLI} to cancer prognosis~\citep{Chowell2021ImprovedPO} are still tree-based ensemble models such as boosting~\citep{Chen2016XGBoostAS,Ke2017LightGBMAH} and bagging~\citep{Breiman2004RandomF,Ishwaran2019RandomSF}. In contrast, transformers~\citep{Vaswani2017AttentionIA}  have  revolutionized enormous fields including natural language processing (NLP)~\citep{Devlin2019BERTPO,Brown2020LanguageMA} and computer vision~\citep{Dosovitskiy2021AnII}. Many attempts to apply transformers on tabular modeling (\emph{e.g.}, TabTransformer~\citealp{Huang2020TabTransformerTD}) have also achieved success. Considering that the disparity between clinical data and other natural tabular data is not large, it is appealing that we can also translate this success from other domains to clinical prediction. As such, we seek to answer the first question in this paper: \emph{To what extent can transformers promote the performance of clinical prediction compared to conventional machine learning approaches?}

Although transformers have advantages in modeling high-dimensional tabular data thanks to the capacity of long-distance dependency modeling, their efficacy can still be hampered when labeled data is scarce given the nature of data-hungry and low inductive bias~\citep{dAscoli2021ConViTIV}. This could be vital to predicting many rare disease areas where historical patient records are extremely limited~\citep{Haendel2019HowMR}. Our pilot investigations (see Figure~\ref{fig:intro}) also confirmed this. Meanwhile, we seek to provide a systematic solution to the clinical prediction that functions both in the presence and absence of much labeled data. Recently, large language models (LLMs) built as a stack of transformers such as BERT~\citep{Devlin2019BERTPO}, GPT-3~\citep{Brown2020LanguageMA} provide a viable direction. The simple and scalable self-supervised learning (\eg masked signal prediction~\citep{Devlin2019BERTPO,Chen2022MaskedIM}) on a nearly unlimited corpus of text (\eg PubMed\footnote{\url{https://pubmed.ncbi.nlm.nih.gov/}}, PMC\footnote{\url{https://www.ncbi.nlm.nih.gov/pmc/}}) has led LLMs to not only continuous performance improvements but also a surprising emergence of in-context learning capability, which is especially powerful under settings with only a small number of learning samples also known as few-shot learning~\citep{Snell2017PrototypicalNF,Sanh2022MultitaskPT}. Though recent work has demonstrated that LLMs are good few-shot clinical information extractors~\citep{Agrawal2022LargeLM}, this success has yet not been extended to tasks with a higher precision requirement, such as cancer prognostic prediction. In this work, we therefore seek to address this second question: \emph{How can language models boost clinical prediction in few-shot settings?}

In addressing these questions, we conduct a benchmarking study on a real-world clinical dataset MSK-IMPACT~\citep{Chowell2021ImprovedPO} to assess the efficacy of a set of baselines and LLMs on prognostic prediction across multiple cancer types (melanoma, NSCLC, bladder, etc.). More importantly, we explore how different pretrained LLMs using different knowledge resources (domain-specific or domain-agnostic) may affect the downstream performance of clinical prediction, especially under few-shot settings. Our results show significant improvements in accuracy through overall survival, progression-free survival and best overall response prediction across multiple disease types.

\section{LLMs for Few-Shot Clinical Prediction}

Figure~\ref{fig:llms} is an overview of applying LLMs for clinical prediction. As discussed in Section~\ref{sec:intro}, purely supervised learning via transformer encoders is often hampered when training samples are limited. LLMs provide a viable direction with astonishing in-context learning capability that exploits knowledge from other resources to downstream tasks with minimal tuning. 
\paragraph{Serialization.}  To leverage LLMs on clinical tabular data, the feature columns must be serialized into sequences of natural language tokens that LLMs can comprehend and encode. Recently, there have been a few trials~\citep{Yin2020TaBERTPF,Bertsimas2022TabTextAS} investigating various serialization techniques and exploring the corresponding performance across different tasks, which turns out that LLMs for tabular modeling rely more on the
correct values than the structure of the features~\citep{Hegselmann2022TabLLMFC}. To avoid repetitive work, in this work, we focus more on how different pretrained LLMs using different knowledge sources may affect the prediction performance by simply following a manual serialization template, \texttt{The \{attribute\} is \{value\}.}, which has been proven to generate competitive results compared to other LLMs prompting-based regeneration methods by~\citet{Hegselmann2022TabLLMFC}.

\paragraph{Knowledge Sources.} The pretraining corpus is also known as the knowledge source for LLMs. Clinical language is notably different from the standard NLP text in terms of vocabulary and syntax~\cite{Wu2019DeepLI}. As a result, following advancements in language modeling from the larger NLP community, the clinical NLP sub-community frequently trains domain-specific models on clinical corpora. Following BERT~\cite{Devlin2019BERTPO}, various clinical and biomedical versions appeared quickly, including BioBERT~\cite{Lee2019BioBERTAP}, ClinicalBERT~\cite{Alsentzer2019PubliclyAC}, SciBERT~\cite{Beltagy2019SciBERTAP}, PubMedBERT~\cite{Gu2020DomainSpecificLM}, etc. However, domain-agnostic LLMs like GPT-3 have so far been unable to achieve competitive results on biomedical NLP tasks~\cite{Moradi2021GPT3MA,Gutierrez2022ThinkingAG}, revealing the fact that the relevance and the knowledge reservation of pretraining sources have a significant impact to the knowledge migration in downstream tasks (\eg finetuning or prompting). Thus, we aim to evaluate the downstream performance in few-shot settings with a few different LLMs pretrained on different resources and benchmark the gaps.

\paragraph{Omnivorous Loss Objective.}
Compared to conventional machine learning approaches, deep learning allows efficient end-to-end learning of image/text encoders in the presence of multi-modality along with tabular data benefiting from the modularized design. More importantly, the customized loss objectives corresponding to different tasks can often be combined for joint training, also known as multi-task learning~\citep{Ruder2017AnOO}. The inductive transfer across related tasks can help improve a model by introducing an inductive bias, which causes a model to prefer some hypotheses over others, that generally leads to solutions that generalize better. In cancer prognostic prediction, we usually have multiple endpoints to predict. For example, \emph{overall survival} (OS), \emph{progression-free survival} (PFS), and \emph{best overall response} (BOR), etc. As such, in this work, we consistently adopt a join learning paradigm that merges multiple endpoints into one unified loss objective $L_{f}$ for all studies using the following term:
\begin{equation}
L_{f} = \sum_{i}^{I}{\alpha_{i} \ell_{i}}
\label{eq:loss}
\end{equation}
where $I$ is the total number of tasks and $\alpha_{i}$ represents the soft weight for any task $i$. More specifically, in our experiments, we adopt CrossEntropy loss for BOR and CoxPH loss for OS and PFS prediction following DeepSurv~\cite{Katzman2018DeepSurvPT}.
% \paragraph{Regression with Finetuning.}

\begin{figure}[!t]	
\centering
\includegraphics[width=0.8\linewidth]{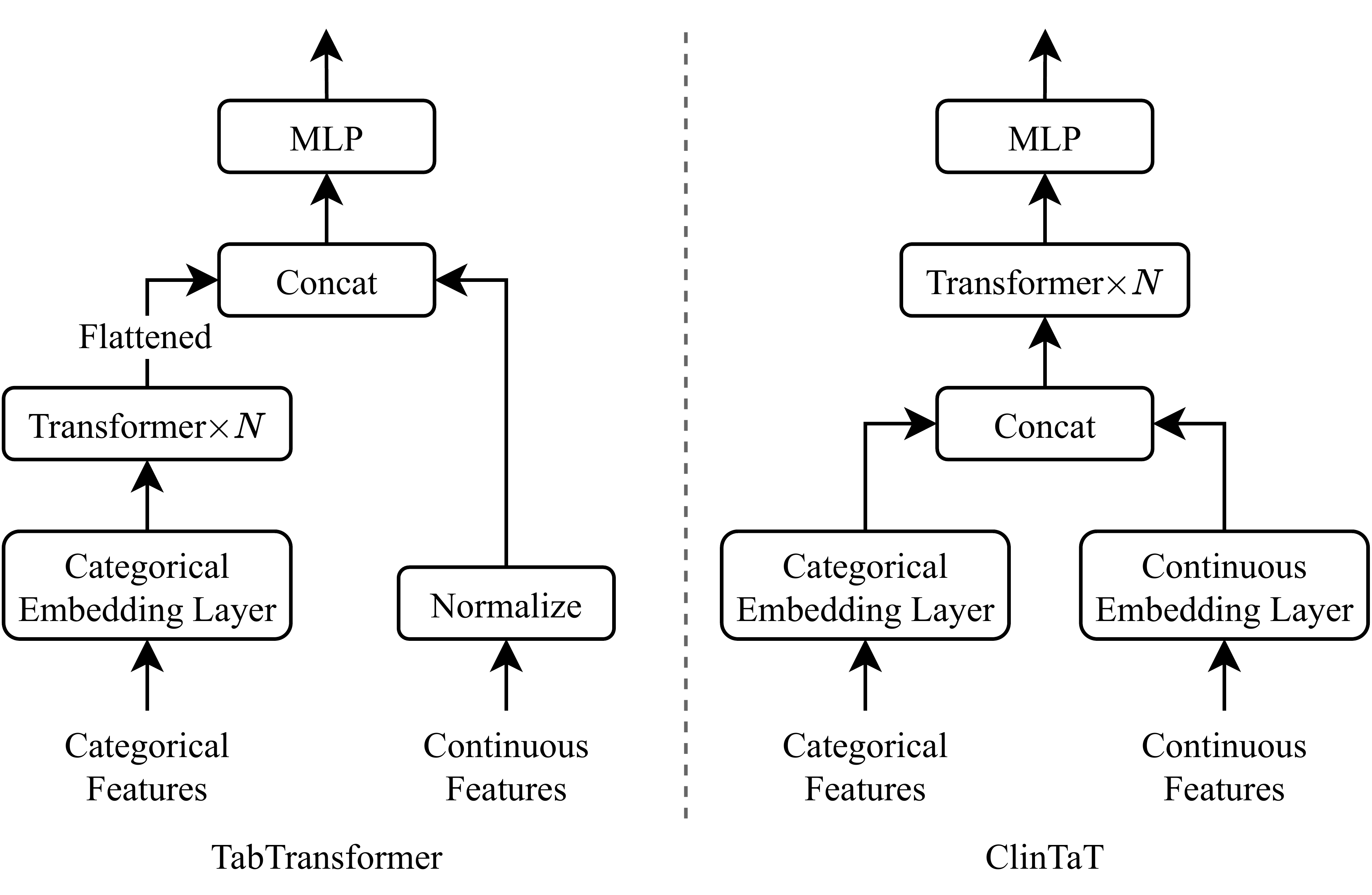}
\caption{\textbf{An illustration of ClinTaT (right).} Compared to original TabTransformer (left), we add a continuous embedding layer for modeling continuous features (\eg lab values) and feed the concatenated inputs into the transformer backbone.}
\label{fig:clintat}
\end{figure}
%\vspace{-10pt}

\begin{figure*}[!t]
\centering
\includegraphics[width=0.24\linewidth]{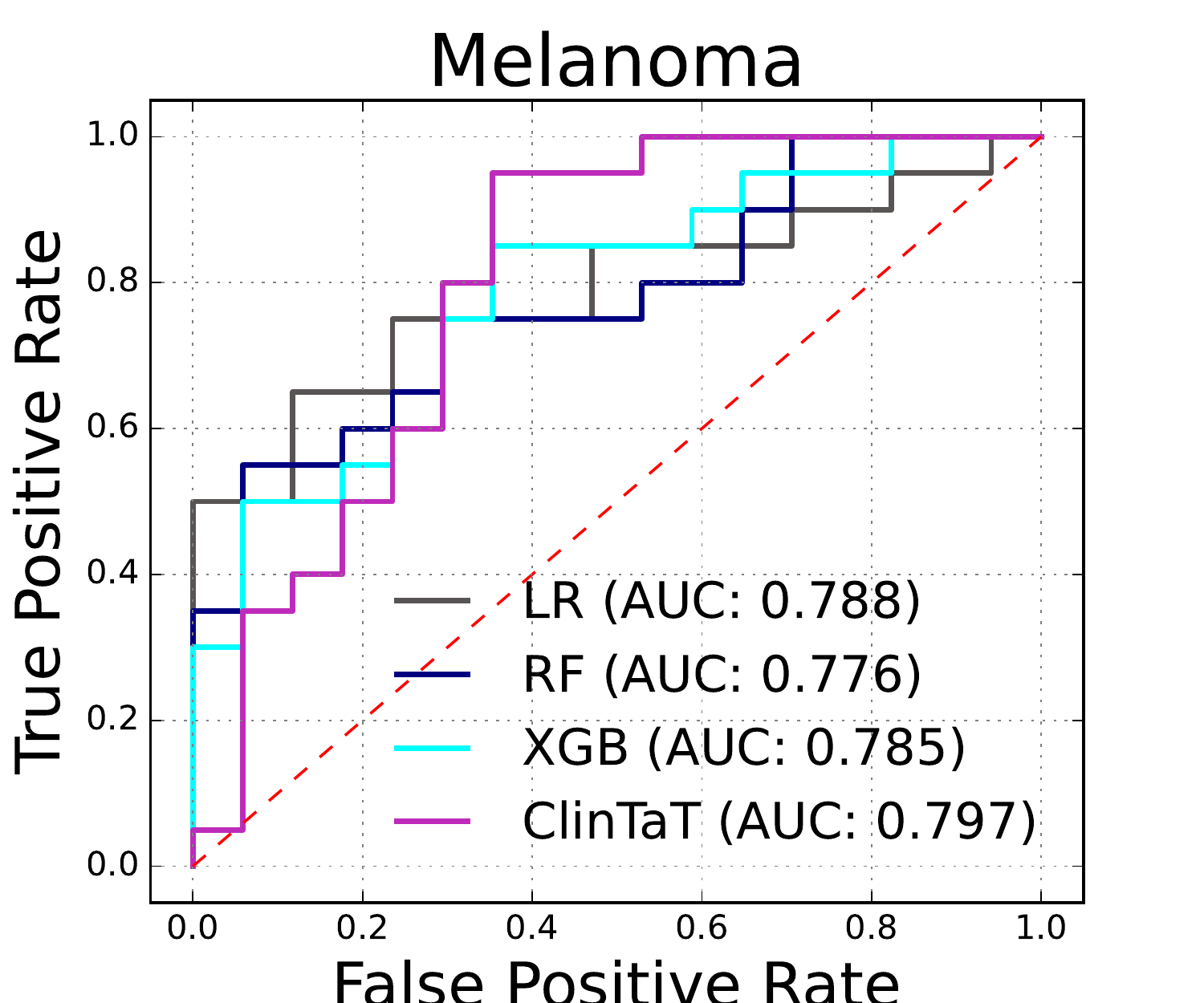}
\includegraphics[width=0.24\linewidth]{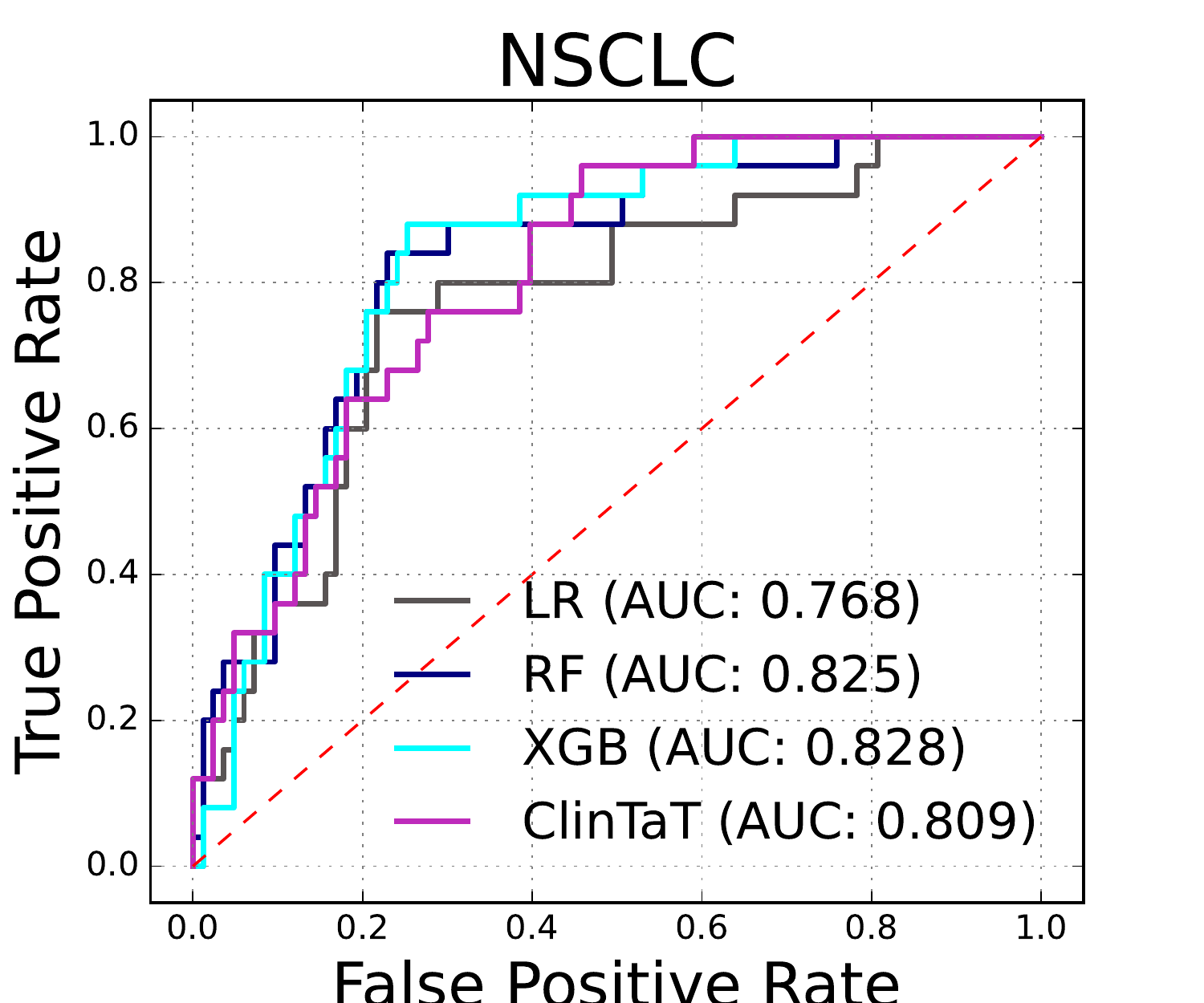}
\includegraphics[width=0.24\linewidth]{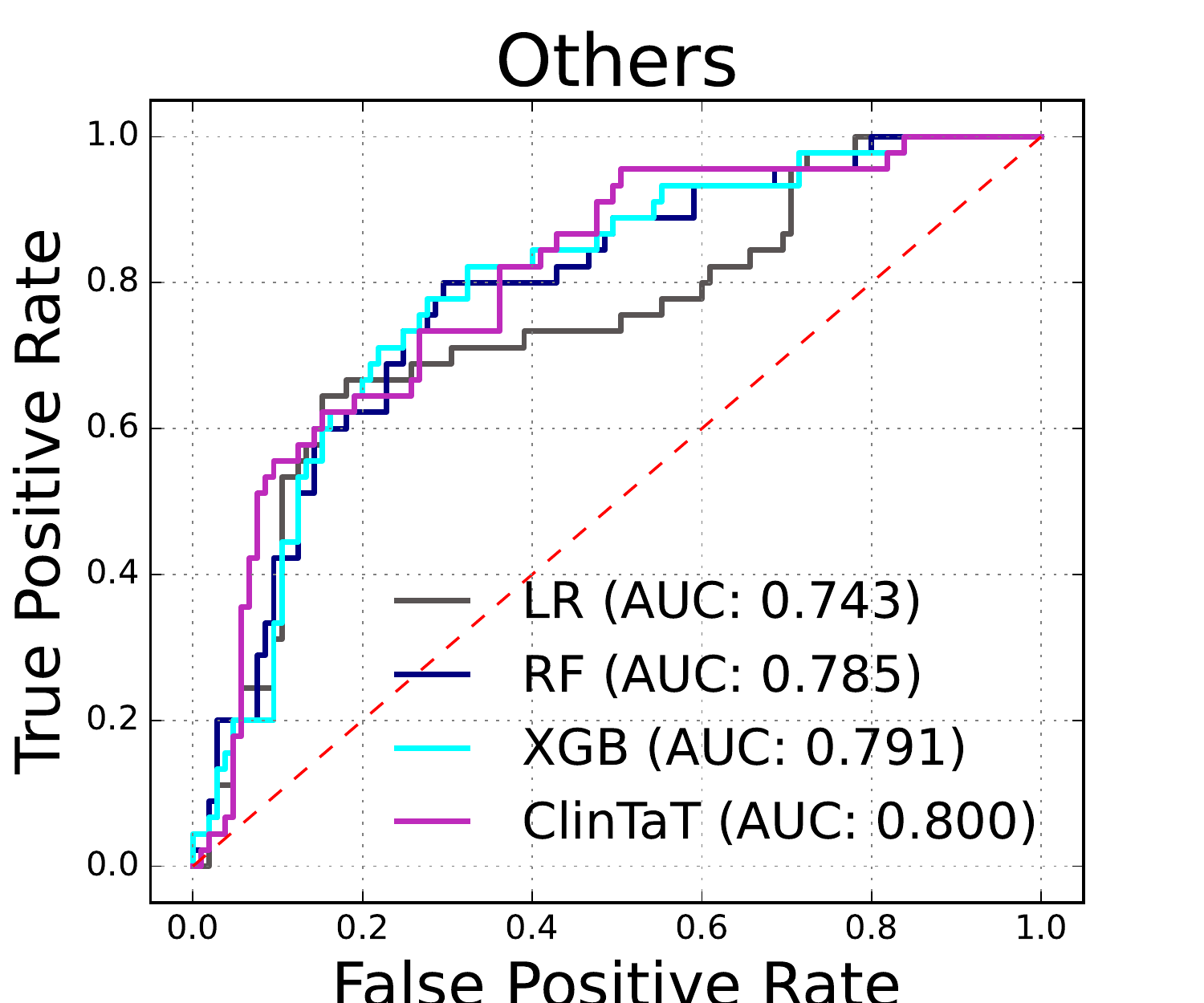}
\includegraphics[width=0.24\linewidth]{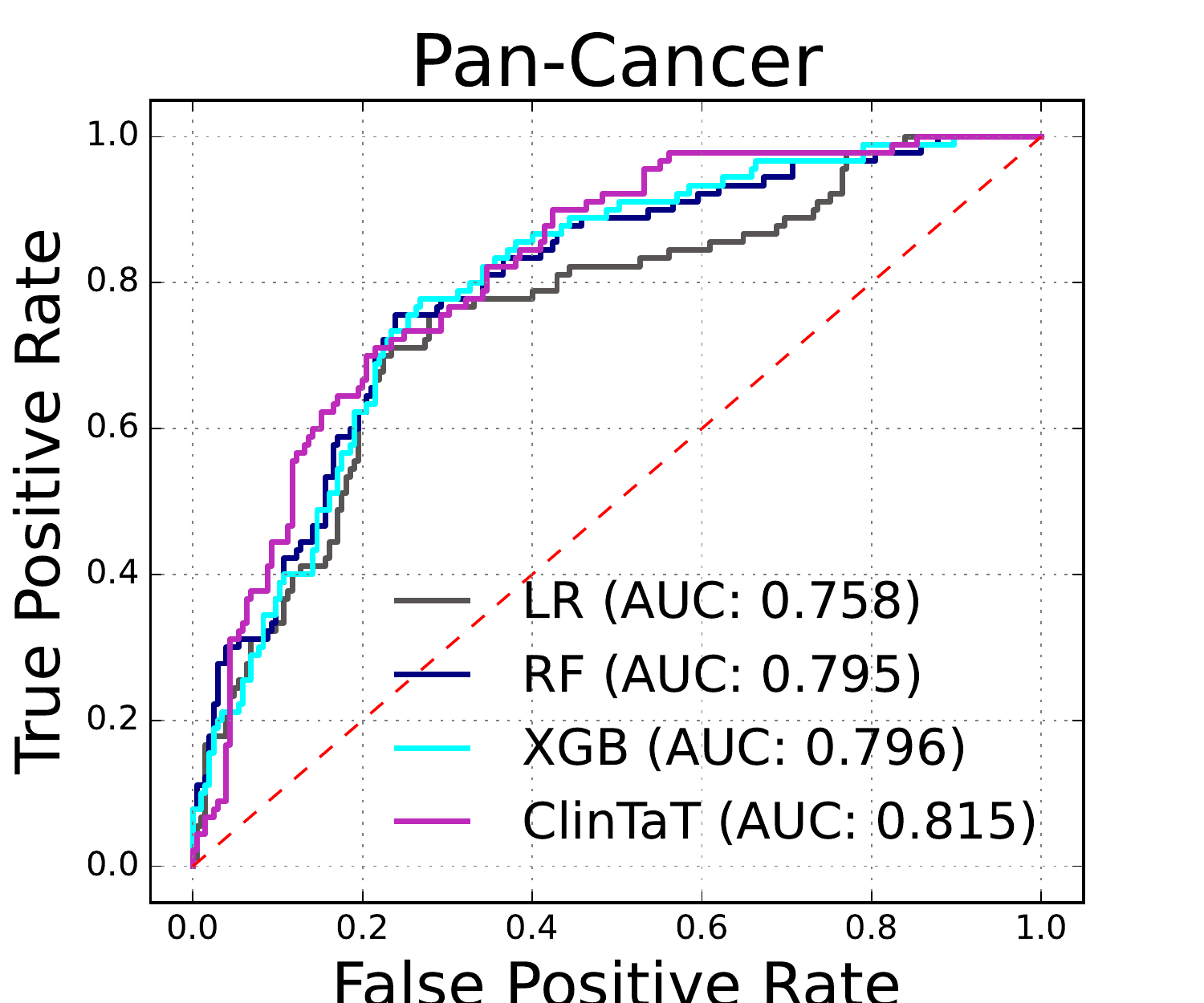}
\caption{\textbf{Model performance across multiple cancer types on test data.} Comparison of predictive performance on MSK-IMPACT in terms of \textbf{ROC} curves and \textbf{AUC} between ClinTaT and other baselines in melanoma, NSCLC, other cancer types and Pan-cancer.}
\label{fig:auc}
\end{figure*}

\begin{figure*}[!t]
\centering
\includegraphics[width=0.3\linewidth]{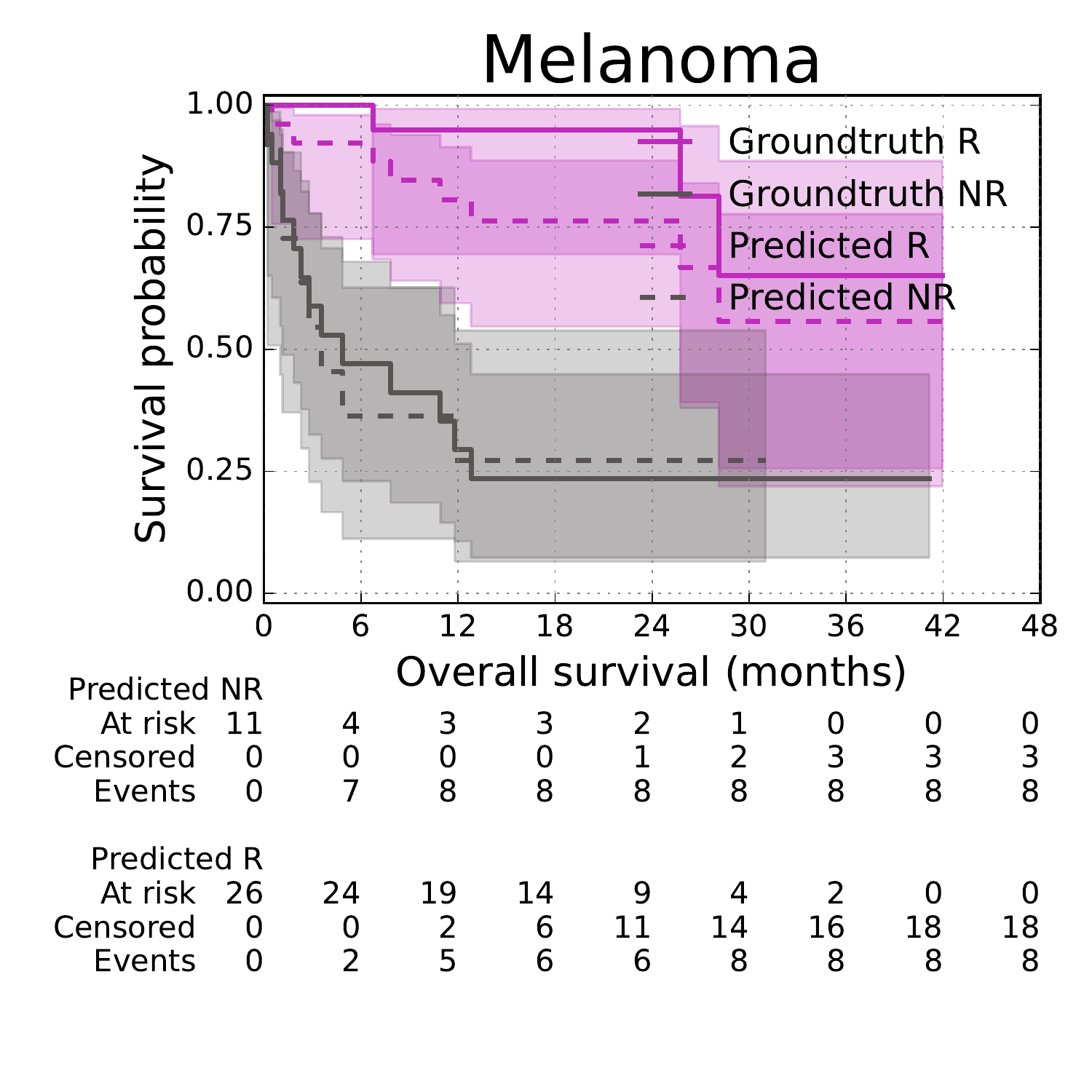}
\includegraphics[width=0.3\linewidth]{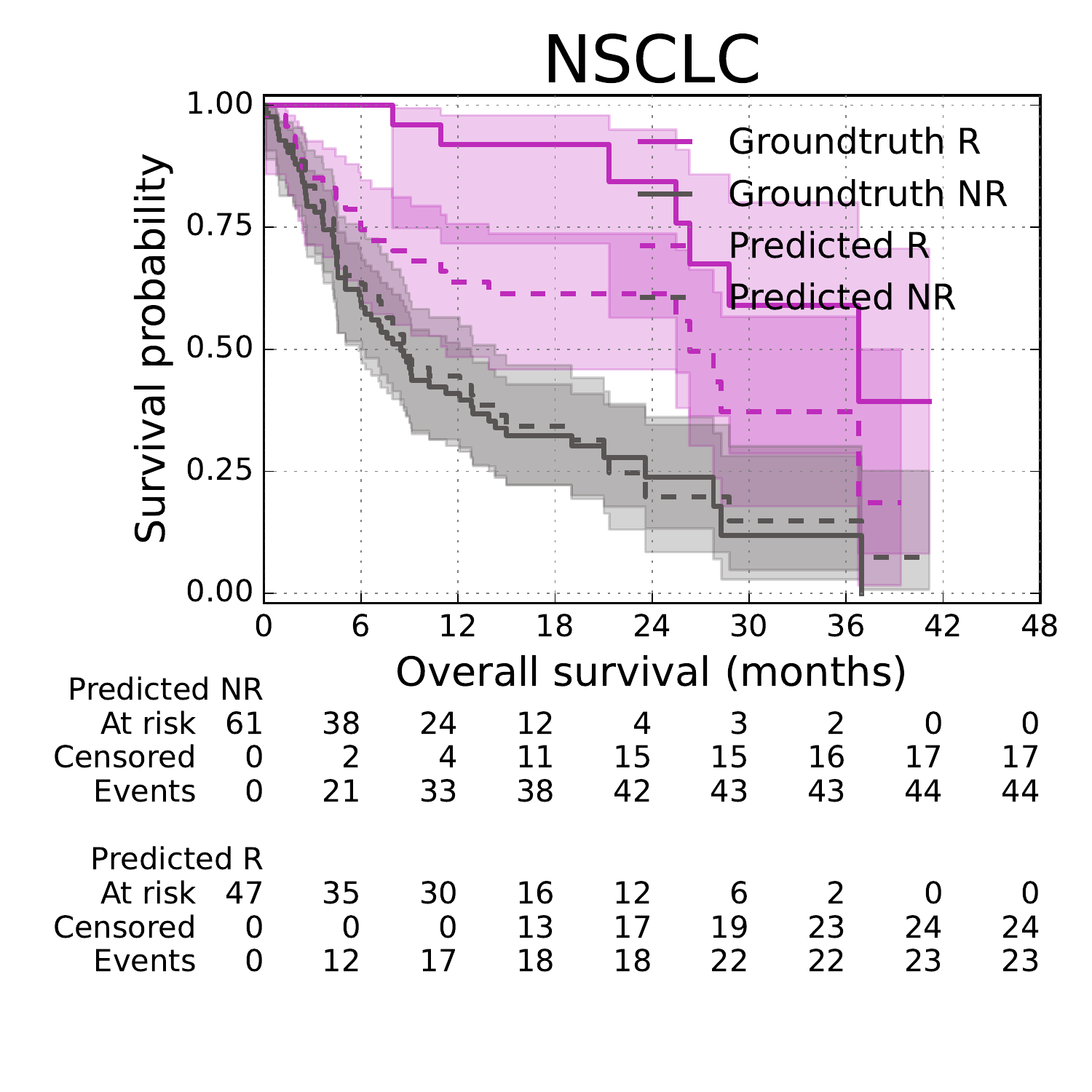}
\includegraphics[width=0.3\linewidth]{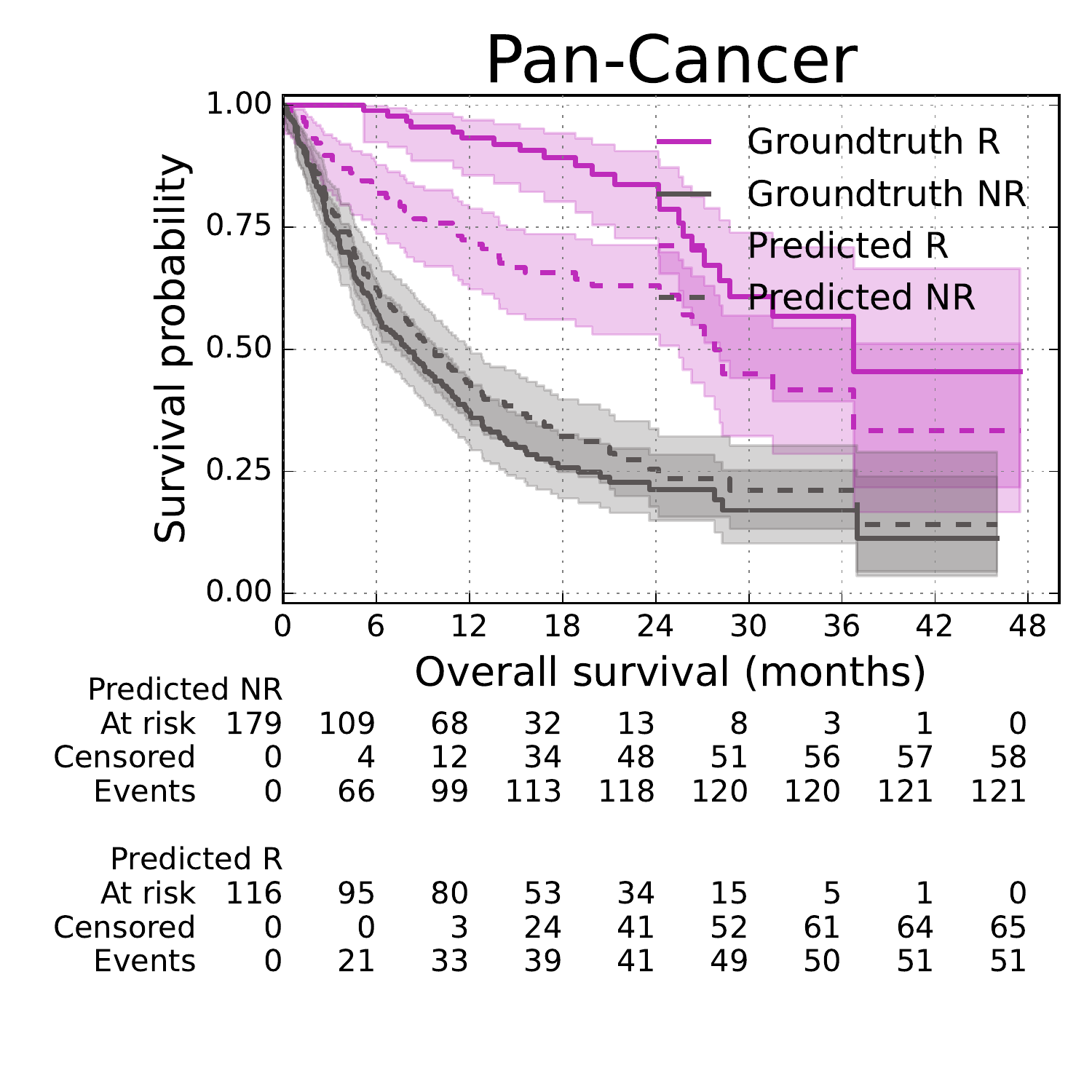}
\caption{\textbf{Model predicts OS and PFS across multiple cancer types on the test data.} Comparison of differences in  \emph{overall survival} between predicted responders and non-responders across multiple cancer types by ClinTaT.}
\label{fig:surv}
\end{figure*}

\section{Experiments and Results}

\paragraph{Data.} This dataset is acquired by Memorial Sloan Kettering Cancer Center (MSKCC) from a comprehensively curated cohort (MSK-IMPACT) with 1,479 patients treated with immune checkpoint blockade (ICB) across 16 different cancer types~\cite{Chowell2021ImprovedPO}, where patients are either responder (R) or non-responders (NR) to the treatment (PD-1/PD-L1 inhibitors, CTLA-4 blockade or a combination) based on Response Evaluation Criteria in Solid Tumors (RECIST) v1.1~\cite{Eisenhauer2009NewRE} or best overall response on imaging. Each patient was collected up to 16 biological features, including genomic, molecular, clinical, and demographic variables. The train set contains 1,184 patients, and the test set contains 295 patients. The evaluation target is to predict \emph{clinical response} to immunotherapy (binary classification) and both \emph{overall survival} and \emph{progression-free survival} (regression) in the test data across different cancer.

\paragraph{Transformers for Tabular Modeling.} As we need to compare with transformer baselines, we also introduce ClinTaT (see Figure~\ref{fig:clintat} right) with some improvements based on the original TabTransformer~\cite{Huang2020TabTransformerTD}, including 1) adding a continuous embedding layer which is consisted of several independent linear layers corresponding to the number of continuous features; 2) directly concatenating the embedded categorical and continuous variables together, and feed them into the transformer instead of only categorical variables. 

\paragraph{Training settings.} For fair comparison, we adopt a hidden dimensionality of $768$ for both ClinTaT and BERTs (base versions). Specifically, ClinTaT is a stack of $6$ transformer encoder layers with $8$ heads. To prevent overfitting, we set the attention dropout rate to $0.3$ and feedforward dropout rate to $0.1$. For BERTs, all layers are frozen while we add one independent encoder on top of it to finetune. In the main figures and tables, we utilize a single linear layer to demonstrate the feasibility of LLMs for few-shot regimes. In ablation studies, we also investigate other encoder types such as another small transformer encoder. The optimizer of AdamW is adopted consistently for all trainings, and the basic learning rates for ClinTaT and BERTs are $1.25e^{-4}$ and $1.25e^{-5}$ with a weight decay of $0.01$, correspondingly. A linear warmup (up to 5 epochs with a total training of 200 epochs) with cosine annealing strategy (warmup learning rate is set to $2.5e^{-7}$) is also applied. For other machine learning baselines, we utilize the grid search to find the optimal hyper-parameters and report the best results. More details can be found in the appedix.

\begin{table*}[!t]
\center
\resizebox{0.75\textwidth}{!}{%
\begin{tabular}{lcccccccccc}
\toprule
\multirow{2}{*}{Model} & \multicolumn{9}{c}{Number of Samples}                                 \\ \cmidrule{2-10} 
                       & 6     & 12    & 18    & 24    & 30    & 36    & 42    & 48    & all \\ \hline
LogRes    & 0.534 & 0.535 & 0.573 & 0.511 & 0.527 & 0.570 & 0.601 & 0.678 & 0.758 \\
RandomForest          & \textbf{0.643} & 0.527 & \textbf{0.672} & 0.539 & 0.594 & \textbf{0.667} & 0.651 & \textbf{0.701} & 0.795 \\
XgBoost                & 0.500 & 0.602 & 0.670 & 0.586 & \textbf{0.613} & 0.664 & 0.651 & 0.681 & 0.796 \\
ClinTaT$_{\texttt{ours}}$                & 0.641 & \textbf{0.619} & 0.653 & \textbf{0.607} & 0.584 & 0.659 & \textbf{0.664} & 0.676 & \textbf{0.815} \\ \bottomrule
\end{tabular}%
}
\caption{Test \textbf{AUC} performance on treatment response prediction of ClinTaT and other baselines on MSK-IMPACT. Each column reports the \emph{k}-shot
performance for different values of \emph{k}. ClinTaT outperforms other traditional approaches with all training samples, however \emph{not significant} in the most few-shot regimes.}
\label{tab:tab1}
\end{table*}
%------------------------------------------------------------------------------------------------------------------------------------------
\begin{table*}[!t]
\center
\resizebox{.75\textwidth}{!}{%
\begin{tabular}{lcccccccccc}
\toprule
\multirow{2}{*}{Model} & \multicolumn{9}{c}{Number of Samples}                                 \\ \cmidrule{2-10} 
                       & 6     & 12    & 18    & 24    & 30    & 36    & 42    & 48    & all \\ \hline
LogRes    & 0.500 & 0.503 & 0.551 & 0.511 & 0.545 & 0.557 & 0.549 & 0.564 & 0.649 \\
RandomForest          & \textbf{0.637} & 0.502 & \textbf{0.614} & 0.536 & 0.591 & 0.610 & 0.626 & 0.631 & 0.682 \\
XgBoost                & 0.500 & 0.555 & 0.601 & 0.539 & \textbf{0.618} & 0.628 & 0.614 & 0.609 & 0.688 \\
ClinTaT$_{\texttt{ours}}$                & 0.583 & \textbf{0.615} & \textbf{0.614} & \textbf{0.639} & 0.610 & \textbf{0.647} & \textbf{0.643} & \textbf{0.645} & \textbf{0.724} \\ \bottomrule
\end{tabular}%
}
\caption{Test \textbf{C-index} performance on \emph{Overall Survival} prediction of ClinTaT and other baselines on MSK-IMPACT. ClinTaT generally outperforms other traditional approaches under many settings, however still \emph{not significant} in the very-few-shot regime (\eg $\le$ 6 samples).}
\label{tab:tab2}
\end{table*}
%---------------------------------------------------------------------------------------------------------------------------------------------
\begin{table*}[!t]
\center
\resizebox{.75\textwidth}{!}{%
\begin{tabular}{lcccccccccc}
\toprule
\multirow{2}{*}{Model} & \multicolumn{9}{c}{Number of Samples}                                 \\ \cmidrule{2-10} 
                       & 6     & 12    & 18    & 24    & 30    & 36    & 42    & 48    & all \\ \hline
LogRes    & 0.515 & 0.513 & 0.538 & 0.514 & 0.537 & 0.549 & 0.565 & 0.596 & 0.648 \\
RandomForest          & \textbf{0.611} & \textbf{0.529} & \textbf{0.612} & 0.532 & 0.580 & 0.619 & \textbf{0.615} & \textbf{0.627} & 0.666 \\
XgBoost                & 0.500 & 0.514 & 0.594 & \textbf{0.569} & \textbf{0.600} & \textbf{0.619} & 0.612 & 0.620 & 0.671 \\
ClinTaT$_{\texttt{ours}}$                & 0.585 & 0.505 & 0.547 & 0.520 & 0.538 & 0.553 & 0.555 & 0.617 & \textbf{0.684} \\ \bottomrule
\end{tabular}%
}
\caption{Test \textbf{C-index} performance on \emph{Progression-free Survival} prediction of ClinTaT and other baselines on MSK-IMPACT. ClinTaT performs better than other approaches only with all training samples.}
\label{tab:tab3}
\end{table*}

%----------------------------------------------------------

\paragraph{How do transformers promote clinical prediction performance?} We first calculated the area under the receiver operating characteristic (ROC) curves using the response probabilities computed by transformers and other baselines. Our proposed ClinTaT achieved superior performance on the test set, as indicated by the area under the curve (AUC), in predicting responders and non-responders across cancer types compared to conventional machine learning models such as logistic regression, random forest, and XgBoost, suggesting that the self-attention mechanism for long-range dependency modeling contributed to the overall prediction performance. (Figure~\ref{fig:auc}, Table~\ref{tab:tab1} using all samples). Furthermore, the differences in OS between responders and non-responders predicted by transformers were significantly higher than differences between responder and non-responder groups predicted by other baselines across various cancer types (Figure~\ref{fig:surv}). Especially for the predicted non-responders, the predicted survival curves almost fit the ground-truth ones perfectly, while it is interesting to observe that transformers tend to underestimate the response probability with an attempt to balance out the prediction performance across different cancer types compared to other baselines ($0.809$ of ClinTaT versus $0.828$ of XGB in Fig.~\ref{fig:auc}). It is additionally beneficial to rare diseases prediction when the training sample pool is not large.

\begin{table*}[!t]
\center
\resizebox{\linewidth}{!}{%
\begin{tabular}{lccccccccc}
\toprule
\multirow{2}{*}{Model} & \multicolumn{8}{c}{Number of Samples}                                  \\ \cmidrule{2-9} 
                       & 4     & 6     & 8              & 10    & 12    & 14    & 16    & 18    \\ \midrule
ClinTaT$_{\texttt{baseline}}$ & 0.593 & 0.641 & 0.638          & 0.628 & 0.619 & 0.643 & 0.639 & 0.653 \\ \midrule
BERT~\citep{Devlin2019BERTPO}  & 0.590  & 0.618 & \textbf{0.652} & 0.636 & 0.633 & 0.637 & 0.632 & 0.631 \\
BioBERT~\citep{Lee2019BioBERTAP}               & 0.570  & 0.512 & 0.527          & 0.532 & 0.536 & 0.532 & 0.524 & 0.530  \\
SciBERT~\citep{Beltagy2019SciBERTAP}               & 0.506 & 0.506 & 0.578          & 0.577 & 0.560  & 0.549 & 0.513 & 0.557 \\
ClinBERT~\citep{Alsentzer2019PubliclyAC}           & 0.604 & 0.550  & 0.545          & 0.560  & 0.567 & 0.576 & 0.574 & 0.558 \\
PubMedBERT~\citep{Gu2020DomainSpecificLM} & \textbf{0.649}$_{\textcolor{purple}{(\uparrow 9.4\%)}}$ & \textbf{0.643}$_{\textcolor{purple}{(\uparrow 0.3\%)}}$ & 0.641$_{\textcolor{purple}{(\uparrow 0.5\%)}}$ & \textbf{0.657}$_{\textcolor{purple}{(\uparrow 4.6\%)}}$ & \textbf{0.663}$_{\textcolor{purple}{(\uparrow 7.1\%)}}$ & \textbf{0.677}$_{\textcolor{purple}{(\uparrow 5.3\%)}}$ & \textbf{0.695}$_{\textcolor{purple}{(\uparrow 8.8\%)}}$ & \textbf{0.685}$_{\textcolor{purple}{(\uparrow 4.9\%)}}$ \\ \bottomrule
\end{tabular}%

}
\caption{Few-shot learning \textbf{AUC} performance of ClinTaT and variants of language models pretrained with different corpus sources on MSK-IMPACT. Best results are in bold and the relative improvements have been marked in \textcolor{purple}{purple}. PubMedBERT~\cite{Gu2020DomainSpecificLM} generally outperforms all the other variants across most settings with an average of improvements over 5\%.}
\label{tab:tab4}
\end{table*}

\comment{
\begin{table}[!t]
\resizebox{\textwidth}{!}{%
\begin{tabular}{lcc}
\toprule
Clinical Terms & BioBERT            & PubMedBERT \\ \midrule
age & $\checkmark$ & $\checkmark$ \\
tumor stage    & $\checkmark$       & $\checkmark$           \\
bmi & b-mi & $\checkmark$ \\
immunotherapy  & im-mu-not-her-ap-y & $\checkmark$           \\
chemotherapy   & ch-em-otherapy     & $\checkmark$           \\
melanoma       & me-lan-oma         & $\checkmark$           \\
nsclc          & n-s-c-l-c          & $\checkmark$           \\
albumin        & album-in           & $\checkmark$           \\
hemologbin     & hem-og-lo-bin      & $\checkmark$           \\
neutrophil     & ne-ut-rop-hil      & $\checkmark$           \\
heterozygosity & he-tero-zy-gos-ity & $\checkmark$           \\ \bottomrule
\end{tabular}%
}
\caption{Comparison of standard clinical terms in vocabularies used by the BioBERT and PubMedBERT. A $\checkmark$ indicates the clinical term appears in the corresponding vocabulary; otherwise, the term will be broken into word pieces as separated by a hyphen. These word pieces often have no biomedical relevance and may hinder learning in downstream tasks.}
\label{tab:tab6}
\end{table}
}

\begin{table}[!t]
\resizebox{\linewidth}{!}{%
\begin{tabular}{llccc}
\toprule
Backbone                      & Encoder     & AUC   & C$_\texttt{OS}$& C$_\texttt{PFS}$ \\ \midrule
\multirow{2}{*}{BERT}         & linear      & 0.725 & 0.593        & 0.622 \\
                              & transformer & \textcolor{purple}{0.773} & \textcolor{purple}{0.699}        & \textcolor{purple}{0.657}         \\
\multirow{2}{*}{BioBERT}      & linear      & 0.678 & 0.590        & 0.625         \\
                              & transformer & \textcolor{purple}{0.766} & \textcolor{purple}{0.707}        & \textcolor{purple}{\textbf{0.672}}         \\
\multirow{2}{*}{SciBERT}      & linear      & 0.689 & 0.588        & 0.620         \\
                              & transformer & \textcolor{purple}{\textbf{0.786}} & \textcolor{purple}{0.711}        & \textcolor{purple}{0.656}         \\
\multirow{2}{*}{ClinBERT} & linear      & 0.669 & 0.591        & 0.616         \\
                              & transformer & \textcolor{purple}{0.751} & \textcolor{purple}{\textbf{0.719}}        & \textcolor{purple}{0.665}         \\
\multirow{2}{*}{PubMedBERT}   & linear      & 0.745 & 0.599        & 0.634         \\
                              & transformer & \textcolor{purple}{0.771} & \textcolor{purple}{0.700}        & \textcolor{purple}{0.662}         \\ \bottomrule
\end{tabular}%
}
\caption{Ablation study on applying different encoders for finetuning of treatment response prediction, including a simple linear layer and a six-layer transformer encoder. Best results across backbones are in bold. Best results across encoders are marked by \textcolor{purple}{purple}. An additional transformer encoder on top of LLMs consistently performs better than a simple linear layer.}
\label{tab:tab5}
\end{table}

To test whether our approach could also predict overall survival (OS) before the administration of immunotherapy, we further calculated the concordance index (C-index) for OS and PFS, which ranges between 0 and 1 (0.5 being random performance). We found that the C-indices of the ClinTaT predictions were significantly higher than those generated by other baselines (Table~\ref{tab:tab2}, pan-cancer C-index $0.724$ for ClinTaT versus $0.688$ for XgBoost versus $0.682$ for Random Forest, $\emph{p}<0.05$; Table~\ref{tab:tab3}, pan-cancer C-index $0.684$ for ClinTaT versus $0.671$ for XgBoost versus $0.666$ for Random Forest, $\emph{p}<0.05$). These results demonstrate that the transformers can accurately forecast response, OS, and PFS before administering immunotherapy.

However, Table~\ref{tab:tab1},~\ref{tab:tab2} and~\ref{tab:tab3} also show that under settings with only a small number of samples, the prediction capability of transformers does not generalize well (\eg $0.583$ for ClinTaT versus $0.637$ for Random Forest with only 6 samples on OS prediction; $0.585$ for ClinTaT versus $0.611$ for Random Forest with only 6 samples on PFS prediction) due to the nature of data-hungry and low inductive bias (discussed in Section~\ref{sec:intro}).

\paragraph{How do LLMs boost few-shot learning?}
Table~\ref{tab:tab4} shows the performance of different BERTs pretrained on different resource corpus followed by a \emph{single linear layer} for finetuning using only \texttt{[cls]} token on MSK-IMPACT test data (averaged over three seeds). The PubMedBERT~\cite{Gu2020DomainSpecificLM} outperforms all other variants and the baseline transformer across all $k$-shot settings with an average of improvements over 5\%. In the very few shot settings (4 samples), the language model finetuning shows significant improvements over the baseline (Table~\ref{tab:tab4}$, 9.4\%$), indicating the benefit of the capability of knowledge transferring to downstream tasks brought by LLMs when samples are insufficient. Also, our results indicate that the sample efficiency of using LLMs' embeddings is highly domain knowledge dependent. The performance of SciBERT is worse than that of BioBERT and ClinicalBERT as SciBERT was pretrained on all semantic scholar 1.14M articles towards a more general scientific knowledge learning.

In contrast, BioBERT and ClinicalBERT were pretrained on the more domain-specific corpus, such as PubMed, PMC, and clinical MIMIC III notes\footnote{\url{https://mimic.mit.edu/}}. However, we cannot claim that domain-specific pretraining is necessary for all clinical prediction tasks as Table~\ref{tab:tab4} also reveals that vanilla BERT is the second best and performs even better than SciBERT pretrained on medical and computer science articles. As we know, vanilla BERT learns more general knowledge understanding from domain-agnostic corpora such as Wikipedia and Book corpus. One of our preliminary conjectures is that domain-specific knowledge transfer is superior when the pretraining corpus is sufficiently profound. However, the generalization capability learned by domain-agnostic models also works under scenarios where the resource knowledge is neither domain-agnostic nor morally domain-specific.

Additionally, the performance down gradation on BioBERT and ClinicalBERT compared to PubMedBERT released more interesting findings as PubMedBERT was pretraining from scratch. At the same time, the other two models were pretrained by inheriting vanilla BERT and BioBERT v1.0\footnote{\url{https://huggingface.co/dmis-lab/biobert-v1.1}}, correspondingly. \citet{Gu2020DomainSpecificLM} has also pointed out that pretraining only sometimes benefits from more text, including out-domain text. The prior biomedical-related BERT models have yet to be pretrained using purely biomedical text. Our Table~\ref{tab:tab4} also shows that domain-specific pretraining from scratch can be superior to mixed-domain pretraining for downstream applications. 

Though all the results in Table~\ref{tab:tab4} are generated by adding one single linear layer on top of LLMs for finetuning, we conduct more ablation studies in Table~\ref{tab:tab5} to evaluate the performance change using different encoders (see Figure~\ref{fig:llms}). The transformer in Table~\ref{tab:tab5} consists of only the transformer encoder of a depth of six layers with a dimension of 768. The results indicate that adding compute complexity to LLMs can still lift the semantic representation learning of clinical features, as transformer architecture performs better than a superficial linear layer. It also provides an alternative way to reexamine the right \emph{size} of LLMs and inspires us for the next step, which is to adopt more scaled LLMs such as PubMedGPT\footnote{\url{https://crfm.stanford.edu/2022/12/15/pubmedgpt.html}}, GPT-3 or T5~\citep{Raffel2019ExploringTL} for clinical prediction.

\section{Limitations}
This study is based on a single clinical cohort consisted of 1479 patients, which may limit the generalizability of the results to other clinical cohorts. This specific cohort of patients may not be representative enough of the general population, which may inject certain level of bias brought by the dissimilar distributions of gender, age, race, etc. While we envision the generalization capability of the language models is applicable to other clinical prediction tasks, the focus of this work is majorly about prognostic prediction of cancer immunotherapy, and we hereby have not provided solid evidence to prove that the success can also be extended to other relevant trials. Additionally, we have yet only compared a limited set of transformers and language models, and it is possible that other models may perform better on the tasks evaluated in this study. Finally, it is important to note that while the models in this study achieve high accuracy in clinical prediction, the ultimate value of these models in improving patient outcomes will depend on how well they are integrated into clinical decision-making processes and the impact they have on patient care. 

\section{Ethical Considerations}
As this work uses real-world patients' clinical data and molecular profiles, which may raise concerns about data privacy and confidentiality. We ensure that all the patients' data is de-identified and protected from unauthorized access and use. The public patient data was approved by the Memorial Sloan Kettering Cancer Center (MSKCC)~\footnote{\url{https://www.mskcc.org/}} institutional review board. Researchers have ensured that they obtain proper ethical approval and informed consent from patients before using their data. Even though this is a dataset that has been carefully curated to prevent the negative impact brought by human bias, there maybe existing a risk of introducing bias into the clinical cohort of data we analyze, particularly in the selection of patients and the choice of clinical features and molecular profiles. Additionally, the use of predictive models to guide clinical decision-making might raise concerns about fair access to healthcare. We hereby ensure that the use of predictive models does not result in the inequitable distribution of healthcare resources and that patients from all socioeconomic backgrounds have equal access to the best possible care. This study uses natural language processing and machine learning algorithms to predict disease prognosis, which may raise broader ethical considerations related to the responsible use of technology in healthcare. We ensure that the use of all approaches discussed in this work is guided by general ethical principles, such as transparency, accountability, and patient-centered care.

Even though we focus on relatively large scale language models in this work, our finetuning strategy only requires a considerably small amount of computation as only the encoder part needs to be finetuned. In practice, the single linear layer finetuning can be obtained in about 2 hours on a machine with single Nvidia A10 GPU; training completes within 5 hours on a machine with one Nvidia A10 GPU for another transformer encoder with a depth of 6 and dimensionality of 768. All the pretrained language model weights are publicly available (\eg~huggingface).

\bibliography{custom}

\begin{thebibliography}{36}
\expandafter\ifx\csname natexlab\endcsname\relax\def\natexlab#1{#1}\fi

\bibitem[{Agrawal et~al.(2022)Agrawal, Hegselmann, Lang, Kim, and
  Sontag}]{Agrawal2022LargeLM}
Monica Agrawal, Stefan Hegselmann, Hunter Lang, Yoon Kim, and David~A. Sontag.
  2022.
\newblock Large language models are zero-shot clinical information extractors.
\newblock \emph{ArXiv}, abs/2205.12689.

\bibitem[{Alsentzer et~al.(2019)Alsentzer, Murphy, Boag, Weng, Jin, Naumann,
  and McDermott}]{Alsentzer2019PubliclyAC}
Emily Alsentzer, John~R. Murphy, Willie Boag, Wei-Hung Weng, Di~Jin, Tristan
  Naumann, and Matthew B.~A. McDermott. 2019.
\newblock Publicly available clinical bert embeddings.
\newblock \emph{ArXiv}, abs/1904.03323.

\bibitem[{Beltagy et~al.(2019)Beltagy, Lo, and Cohan}]{Beltagy2019SciBERTAP}
Iz~Beltagy, Kyle Lo, and Arman Cohan. 2019.
\newblock Scibert: A pretrained language model for scientific text.
\newblock \emph{EMNLP}.

\bibitem[{Bertsimas et~al.(2022)Bertsimas, Carballo, Ma, Na, Boussioux, Zeng,
  Soenksen, and Fuentes}]{Bertsimas2022TabTextAS}
Dimitris Bertsimas, Kimberly~Villalobos Carballo, Yu~Ma, Liangyuan Na,
  L{\'e}onard Boussioux, Cynthia Zeng, Luis~R. Soenksen, and Ignacio Fuentes.
  2022.
\newblock Tabtext: a systematic approach to aggregate knowledge across tabular
  data structures.
\newblock \emph{ArXiv}, abs/2206.10381.

\bibitem[{Breiman(2004)}]{Breiman2004RandomF}
L.~Breiman. 2004.
\newblock Random forests.
\newblock \emph{Machine Learning}, 45:5--32.

\bibitem[{Brown et~al.(2020)Brown, Mann, Ryder, Subbiah, Kaplan, Dhariwal,
  Neelakantan, Shyam, Sastry, Askell, Agarwal, Herbert-Voss, Krueger, Henighan,
  Child, Ramesh, Ziegler, Wu, Winter, Hesse, Chen, Sigler, Litwin, Gray, Chess,
  Clark, Berner, McCandlish, Radford, Sutskever, and
  Amodei}]{Brown2020LanguageMA}
Tom~B. Brown, Benjamin Mann, Nick Ryder, Melanie Subbiah, Jared Kaplan,
  Prafulla Dhariwal, Arvind Neelakantan, Pranav Shyam, Girish Sastry, Amanda
  Askell, Sandhini Agarwal, Ariel Herbert-Voss, Gretchen Krueger, T.~J.
  Henighan, Rewon Child, Aditya Ramesh, Daniel~M. Ziegler, Jeff Wu, Clemens
  Winter, Christopher Hesse, Mark Chen, Eric Sigler, Mateusz Litwin, Scott
  Gray, Benjamin Chess, Jack Clark, Christopher Berner, Sam McCandlish, Alec
  Radford, Ilya Sutskever, and Dario Amodei. 2020.
\newblock Language models are few-shot learners.
\newblock \emph{NeurIPS}, abs/2005.14165.

\bibitem[{Chen and Guestrin(2016)}]{Chen2016XGBoostAS}
Tianqi Chen and Carlos Guestrin. 2016.
\newblock Xgboost: A scalable tree boosting system.
\newblock \emph{SIGKDD}.

\bibitem[{Chen et~al.(2022)Chen, Agarwal, Aggarwal, Safta, Balan, Sethuraman,
  and Brown}]{Chen2022MaskedIM}
Zekai Chen, Devansh Agarwal, Kshitij Aggarwal, Wiem Safta, Mariann~Micsinai
  Balan, Venkat~S. Sethuraman, and Kevin Brown. 2022.
\newblock Masked image modeling advances 3d medical image analysis.
\newblock \emph{WACV}, abs/2204.11716.

\bibitem[{Chowell et~al.(2021)Chowell, Yoo, Valero, Pastore, Krishna, Lee,
  Hoen, Shi, Kelly, Patel, Makarov, Ma, Vuong, Sabio, Weiss, Kuo, Lenz,
  Samstein, Riaz, Adusumilli, Balachandran, Plitas, Hakimi, Abdel-Wahab,
  Shoushtari, Postow, Motzer, Ladanyi, Zehir, Berger, G{\"o}nen, Morris,
  Weinhold, and Chan}]{Chowell2021ImprovedPO}
Diego Chowell, Seong-Keun Yoo, Cristina Valero, Alessandro Pastore, Chirag
  Krishna, Mark Lee, Douglas~R. Hoen, Hongyu Shi, Daniel~W. Kelly, Neal Patel,
  Vladimir Makarov, Xiaoxiao Ma, Lynda Vuong, Erich Sabio, Kate Weiss, Fengshen
  Kuo, Tobias~L. Lenz, Robert~M. Samstein, Nadeem Riaz, Prasad~S. Adusumilli,
  Vinod~P. Balachandran, George Plitas, A.~Ari Hakimi, Omar Abdel-Wahab,
  Alexander~N. Shoushtari, Michael~A. Postow, R.~Motzer, Marc Ladanyi, Ahmet
  Zehir, Michael~F. Berger, Mithat G{\"o}nen, Luc G.~T. Morris, Nils Weinhold,
  and Timothy~A. Chan. 2021.
\newblock Improved prediction of immune checkpoint blockade efficacy across
  multiple cancer types.
\newblock \emph{Nature biotechnology}.

\bibitem[{d'Ascoli et~al.(2021)d'Ascoli, Touvron, Leavitt, Morcos, Biroli, and
  Sagun}]{dAscoli2021ConViTIV}
St{\'e}phane d'Ascoli, Hugo Touvron, Matthew~L. Leavitt, Ari~S. Morcos, Giulio
  Biroli, and Levent Sagun. 2021.
\newblock Convit: improving vision transformers with soft convolutional
  inductive biases.
\newblock \emph{ICLR}, 2022.

\bibitem[{Devlin et~al.(2019)Devlin, Chang, Lee, and
  Toutanova}]{Devlin2019BERTPO}
Jacob Devlin, Ming-Wei Chang, Kenton Lee, and Kristina Toutanova. 2019.
\newblock Bert: Pre-training of deep bidirectional transformers for language
  understanding.
\newblock \emph{NAACL}, abs/1810.04805.

\bibitem[{Dosovitskiy et~al.(2021)Dosovitskiy, Beyer, Kolesnikov, Weissenborn,
  Zhai, Unterthiner, Dehghani, Minderer, Heigold, Gelly, Uszkoreit, and
  Houlsby}]{Dosovitskiy2021AnII}
Alexey Dosovitskiy, Lucas Beyer, Alexander Kolesnikov, Dirk Weissenborn,
  Xiaohua Zhai, Thomas Unterthiner, Mostafa Dehghani, Matthias Minderer, Georg
  Heigold, Sylvain Gelly, Jakob Uszkoreit, and Neil Houlsby. 2021.
\newblock An image is worth 16x16 words: Transformers for image recognition at
  scale.
\newblock \emph{ICLR}, abs/2010.11929.

\bibitem[{Eisenhauer et~al.(2009)Eisenhauer, Therasse, Bogaerts, Schwartz,
  Sargent, Ford, Dancey, Arbuck, Gwyther, Mooney, Rubinstein, Shankar, Dodd,
  Kaplan, Lacombe, and Verweij}]{Eisenhauer2009NewRE}
E.~A. Eisenhauer, Patrick Therasse, Jan Bogaerts, Lawrence~H. Schwartz,
  Daniel~J. Sargent, Robert Ford, Janet~E. Dancey, Susan~G. Arbuck, S.~Gwyther,
  Margaret Mooney, Larry~V. Rubinstein, Lalitha~K Shankar, Lori~E. Dodd,
  Richard~S. Kaplan, Denis Lacombe, and Jaap Verweij. 2009.
\newblock New response evaluation criteria in solid tumours: revised recist
  guideline (version 1.1).
\newblock \emph{European journal of cancer}, 45 2:228--47.

\bibitem[{Gu et~al.(2020)Gu, Tinn, Cheng, Lucas, Usuyama, Liu, Naumann, Gao,
  and Poon}]{Gu2020DomainSpecificLM}
Yuxian Gu, Robert Tinn, Hao Cheng, Michael~R. Lucas, Naoto Usuyama, Xiaodong
  Liu, Tristan Naumann, Jianfeng Gao, and Hoifung Poon. 2020.
\newblock Domain-specific language model pretraining for biomedical natural
  language processing.
\newblock \emph{ACM Transactions on Computing for Healthcare (HEALTH)}, 3:1 --
  23.

\bibitem[{Gutierrez et~al.(2022)Gutierrez, McNeal, Washington, Chen, Li, Sun,
  and Su}]{Gutierrez2022ThinkingAG}
Bernal~Jimenez Gutierrez, Nikolas McNeal, Clay Washington, You Chen, Lang Li,
  Huan Sun, and Yu~Su. 2022.
\newblock Thinking about gpt-3 in-context learning for biomedical ie? think
  again.
\newblock \emph{ArXiv}, abs/2203.08410.

\bibitem[{Haendel et~al.(2019)Haendel, Vasilevsky, Unni, Bologa, Harris, Rehm,
  Hamosh, Baynam, Groza, McMurry, Dawkins, Rath, Thaxon, Bocci, marcin~p.
  joachimiak, K{\"o}hler, Robinson, Mungall, and Oprea}]{Haendel2019HowMR}
Melissa~A. Haendel, N~Vasilevsky, Deepak~R. Unni, Cristian~G Bologa, Nomi~L.
  Harris, Heidi~L. Rehm, Ada Hamosh, Gareth~S. Baynam, Tudor Groza, Julie~A.
  McMurry, Hugh J.~S. Dawkins, Ana Rath, Courtney Thaxon, Giovanni Bocci,
  marcin~p. joachimiak, Sebastian K{\"o}hler, Peter~N. Robinson, Chris~J.
  Mungall, and Tudor~I. Oprea. 2019.
\newblock How many rare diseases are there?
\newblock \emph{Nature Reviews Drug Discovery}, 19:77--78.

\bibitem[{Hegselmann et~al.(2022)Hegselmann, Buendia, Lang, Agrawal, Jiang, and
  Sontag}]{Hegselmann2022TabLLMFC}
Stefan Hegselmann, Alejandro Buendia, Hunter Lang, Monica Agrawal, Xiaoyi
  Jiang, and David~A. Sontag. 2022.
\newblock Tabllm: Few-shot classification of tabular data with large language
  models.
\newblock \emph{ArXiv}, abs/2210.10723.

\bibitem[{Huang et~al.(2020)Huang, Khetan, Cvitkovic, and
  Karnin}]{Huang2020TabTransformerTD}
Xin Huang, Ashish Khetan, Milan~W. Cvitkovic, and Zohar~S. Karnin. 2020.
\newblock Tabtransformer: Tabular data modeling using contextual embeddings.
\newblock \emph{ArXiv}, abs/2012.06678.

\bibitem[{Ishwaran et~al.(2019)Ishwaran, Kogalur, Blackstone, and
  Lauer}]{Ishwaran2019RandomSF}
Hemant Ishwaran, Udaya~B. Kogalur, Eugene~H. Blackstone, and Michael~S. Lauer.
  2019.
\newblock Random survival forests.
\newblock \emph{Wiley StatsRef: Statistics Reference Online}.

\bibitem[{Katzman et~al.(2018)Katzman, Shaham, Cloninger, Bates, Jiang, and
  Kluger}]{Katzman2018DeepSurvPT}
Jared Katzman, Uri Shaham, Alexander Cloninger, Jonathan Bates, Tingting Jiang,
  and Yuval Kluger. 2018.
\newblock Deepsurv: personalized treatment recommender system using a cox
  proportional hazards deep neural network.
\newblock \emph{BMC Medical Research Methodology}, 18.

\bibitem[{Ke et~al.(2017)Ke, Meng, Finley, Wang, Chen, Ma, Ye, and
  Liu}]{Ke2017LightGBMAH}
Guolin Ke, Qi~Meng, Thomas Finley, Taifeng Wang, Wei Chen, Weidong Ma, Qiwei
  Ye, and Tie-Yan Liu. 2017.
\newblock Lightgbm: A highly efficient gradient boosting decision tree.
\newblock In \emph{NeurIPS}.

\bibitem[{Lee et~al.(2019)Lee, Yoon, Kim, Kim, Kim, So, and
  Kang}]{Lee2019BioBERTAP}
Jinhyuk Lee, Wonjin Yoon, Sungdong Kim, Donghyeon Kim, Sunkyu Kim, Chan~Ho So,
  and Jaewoo Kang. 2019.
\newblock Biobert: a pre-trained biomedical language representation model for
  biomedical text mining.
\newblock \emph{Bioinformatics}, 36:1234 -- 1240.

\bibitem[{Moradi et~al.(2021)Moradi, Blagec, Haberl, and
  Samwald}]{Moradi2021GPT3MA}
Milad Moradi, Kathrin Blagec, Florian Haberl, and Matthias Samwald. 2021.
\newblock Gpt-3 models are poor few-shot learners in the biomedical domain.
\newblock \emph{ArXiv}, abs/2109.02555.

\bibitem[{Pardoll(2012)}]{Pardoll2012TheBO}
Drew~M. Pardoll. 2012.
\newblock The blockade of immune checkpoints in cancer immunotherapy.
\newblock \emph{Nature Reviews Cancer}, 12:252--264.

\bibitem[{Raffel et~al.(2019)Raffel, Shazeer, Roberts, Lee, Narang, Matena,
  Zhou, Li, and Liu}]{Raffel2019ExploringTL}
Colin Raffel, Noam~M. Shazeer, Adam Roberts, Katherine Lee, Sharan Narang,
  Michael Matena, Yanqi Zhou, Wei Li, and Peter~J. Liu. 2019.
\newblock Exploring the limits of transfer learning with a unified text-to-text
  transformer.
\newblock \emph{ArXiv}, abs/1910.10683.

\bibitem[{Rajkomar et~al.(2019)Rajkomar, Dean, and
  Kohane}]{Rajkomar2019MachineLI}
Alvin Rajkomar, Jeffrey Dean, and Isaac~S. Kohane. 2019.
\newblock Machine learning in medicine.
\newblock \emph{The New England Journal of Medicine}, 380:1347–1358.

\bibitem[{Ruder(2017)}]{Ruder2017AnOO}
Sebastian Ruder. 2017.
\newblock An overview of multi-task learning in deep neural networks.
\newblock \emph{ArXiv}, abs/1706.05098.

\bibitem[{Sanh et~al.(2022)Sanh, Webson, Raffel, Bach, Sutawika, Alyafeai,
  Chaffin, Stiegler, Scao, Raja, Dey, Bari, Xu, Thakker, Sharma, Szczechla,
  Kim, Chhablani, Nayak, Datta, Chang, Jiang, Wang, Manica, Shen, Yong, Pandey,
  Bawden, Wang, Neeraj, Rozen, Sharma, Santilli, F{\'e}vry, Fries, Teehan,
  Biderman, Gao, Bers, Wolf, and Rush}]{Sanh2022MultitaskPT}
Victor Sanh, Albert Webson, Colin Raffel, Stephen~H. Bach, Lintang Sutawika,
  Zaid Alyafeai, Antoine Chaffin, Arnaud Stiegler, Teven~Le Scao, Arun Raja,
  Manan Dey, M~Saiful Bari, Canwen Xu, Urmish Thakker, Shanya Sharma, Eliza
  Szczechla, Taewoon Kim, Gunjan Chhablani, Nihal~V. Nayak, Debajyoti Datta,
  Jonathan Chang, Mike Tian-Jian Jiang, Han Wang, Matteo Manica, Sheng Shen,
  Zheng~Xin Yong, Harshit Pandey, Rachel Bawden, Thomas Wang, Trishala Neeraj,
  Jos Rozen, Abheesht Sharma, Andrea Santilli, Thibault F{\'e}vry, Jason~Alan
  Fries, Ryan Teehan, Stella~Rose Biderman, Leo Gao, Tali Bers, Thomas Wolf,
  and Alexander~M. Rush. 2022.
\newblock Multitask prompted training enables zero-shot task generalization.
\newblock \emph{ICLR}, abs/2110.08207.

\bibitem[{Smeden et~al.(2021)Smeden, Reitsma, Riley, Collins, and
  Moons}]{vanSmeden2021ClinicalPM}
Maarten~Van Smeden, Johannes~B. Reitsma, Richard~D. Riley, Gary~Stephen
  Collins, and Karel G.~M. Moons. 2021.
\newblock Clinical prediction models: diagnosis versus prognosis.
\newblock \emph{Journal of clinical epidemiology}, 132:142--145.

\bibitem[{Snell et~al.(2017)Snell, Swersky, and
  Zemel}]{Snell2017PrototypicalNF}
Jake Snell, Kevin Swersky, and Richard~S. Zemel. 2017.
\newblock Prototypical networks for few-shot learning.
\newblock \emph{NeurIPS}, abs/1703.05175.

\bibitem[{Steyerberg(2008)}]{Steyerberg2008ClinicalPM}
Ewout~Willem Steyerberg. 2008.
\newblock Clinical prediction models: A practical approach to development,
  validation, and updating.
\newblock In \emph{Springer}.

\bibitem[{Topalian et~al.(2016)Topalian, Taube, Anders, and
  Pardoll}]{Topalian2016MechanismdrivenBT}
Suzanne~L. Topalian, Janis~M. Taube, Robert~A Anders, and Drew~M. Pardoll.
  2016.
\newblock Mechanism-driven biomarkers to guide immune checkpoint blockade in
  cancer therapy.
\newblock \emph{Nature Reviews Cancer}, 16:275--287.

\bibitem[{Topol(2019)}]{Topol2019HighperformanceMT}
Eric~J. Topol. 2019.
\newblock High-performance medicine: the convergence of human and artificial
  intelligence.
\newblock \emph{Nature Medicine}, 25:44--56.

\bibitem[{Vaswani et~al.(2017)Vaswani, Shazeer, Parmar, Uszkoreit, Jones,
  Gomez, Kaiser, and Polosukhin}]{Vaswani2017AttentionIA}
Ashish Vaswani, Noam~M. Shazeer, Niki Parmar, Jakob Uszkoreit, Llion Jones,
  Aidan~N. Gomez, Lukasz Kaiser, and Illia Polosukhin. 2017.
\newblock Attention is all you need.
\newblock \emph{NeurIPS}, abs/1706.03762.

\bibitem[{Wu et~al.(2019)Wu, Roberts, Datta, Du, Ji, Si, Soni, Wang, Wei,
  Xiang, Zhao, and Xu}]{Wu2019DeepLI}
Stephen~T Wu, Kirk Roberts, Surabhi Datta, Jingcheng Du, Zongcheng Ji, Yuqi Si,
  Sarvesh Soni, Qiong Wang, Qiang Wei, Yang Xiang, Bo~Zhao, and Hua Xu. 2019.
\newblock Deep learning in clinical natural language processing: a methodical
  review.
\newblock \emph{Journal of the American Medical Informatics Association :
  JAMIA}.

\bibitem[{Yin et~al.(2020)Yin, Neubig, tau Yih, and Riedel}]{Yin2020TaBERTPF}
Pengcheng Yin, Graham Neubig, Wen tau Yih, and Sebastian Riedel. 2020.
\newblock Tabert: Pretraining for joint understanding of textual and tabular
  data.
\newblock \emph{ACL}, abs/2005.08314.

\end{thebibliography}
\bibliographystyle{acl_natbib}

\appendix

% \section{Example Appendix}
% \label{sec:appendix}

% This is a section in the appendix.

\end{document}